\documentclass[sn-apa]{sn-jnl}

\usepackage{graphicx}
\usepackage{multirow}
\usepackage{amsmath,amssymb,amsfonts}
\usepackage{amsthm}
\usepackage{mathrsfs}
\usepackage[title]{appendix}
\usepackage{xcolor}
\usepackage{textcomp}
\usepackage{manyfoot}
\usepackage{booktabs}
\usepackage{algorithm}
\usepackage{algorithmicx}
\usepackage{algpseudocode}
\usepackage{listings}

\usepackage{lmodern}

\begin{document}

\title[Article Title (short version)]{Understanding and Benchmarking Artificial Intelligence: OpenAI's o3 Is Not AGI}

\author[1, 2]{\fnm{Rolf} \sur{Pfister} \email{rolf.pfister@posteo.de}}
\author[1]{\fnm{Hansueli} \sur{Jud} \email{hansueli.jud@lab42.global}}

\affil[1]{\orgname{Lab42 AI Research Institute}, \orgaddress{\street{Obere Strasse 22b}, \city{Davos}, \postcode{7270}, \country{Switzerland}}}
\affil[2]{\orgdiv{Munich Center for Mathematical Philosophy (MCMP)}, \orgname{Ludwig-Maximilians-Universit\"at M\"unchen}, \orgaddress{\street{Geschwister-Scholl-Platz 1}, \city{Munich}, \postcode{80539}, \country{Germany}}}

\abstract{
OpenAI's o3 achieves a high score of 87.5 \% on ARC-AGI, a benchmark proposed to measure intelligence.
This raises the question whether systems based on Large Language Models (LLMs), particularly o3, demonstrate intelligence and progress towards artificial general intelligence (AGI).
Building on the distinction between skills and intelligence made by François Chollet, the creator of ARC-AGI, a new understanding of intelligence is introduced:
an agent is the more intelligent, the more efficiently it can achieve the more diverse goals in the more diverse worlds with the less knowledge.
An analysis of the ARC-AGI benchmark shows that its tasks represent a very specific type of problem that can be solved by massive trialling of combinations of predefined operations.
This method is also applied by o3, achieving its high score through the extensive use of computing power.
However, for most problems in the physical world and in the human domain, solutions cannot be tested in advance and predefined operations are not available.
Consequently, massive trialling of predefined operations, as o3 does, cannot be a basis for AGI -- instead, new approaches are required that can reliably solve a wide variety of problems without existing skills.
To support this development, a new benchmark for intelligence is outlined that covers a much higher diversity of unknown tasks to be solved, thus enabling a comprehensive assessment of intelligence and of progress towards AGI.
}


\maketitle

\section{Introduction} \label{sec1}

The release of systems based on large language models (LLMs)\footnote{
OpenAI's o3, but also its predecessors and comparable systems from other companies such as Google's Gemini are based on LLMs at their core, but contain many additional modules that improve and extend their functions.
}, in particular \mbox{OpenAI's} ChatGPT in 2022, caused intense and ongoing debates about the extent of their intelligence.
For example, Microsoft, one of the stakeholders in OpenAI, stated that the successor model "GPT-4 attains a form of general intelligence, indeed showing sparks of artificial general intelligence" \citep[p. 92]{817}.
Further statements that LLM-based systems represent artificial general intelligence (AGI) or at least major progress towards it have been made by OpenAI and other prominent AI companies, but also by AI experts, and in the media.
At the same time, others take a more critical perspective on the performance of LLM-based systems, attributing their success not to intelligence, but to other factors such as the vast amounts of training data and the extensive computing resources used.
The discussion has recently intensified again with the success of OpenAI's latest model o3 on the ARC-AGI benchmark, where it achieved 87.5 \% on the semi-private test set; an achievement \citet{arcprize-o3} calls 'a genuine breakthrough, marking a qualitative shift in AI capabilities'.

ARC-AGI, originally called Abstraction and Reasoning Corpus (ARC), is designed as a benchmark for measuring general intelligence and was developed by \citet[pp. 46-58]{674}.
In contrast to other benchmarks, ARC-AGI is not intended to measure the performance of an AI approach in a specific skill, but instead its ability to solve new, unknown tasks which it has not encountered before.
ARC-AGI consists of 1,000 unique tasks, of which 800 are publicly accessible and divided into 400 training tasks and 400 evaluation tasks.
The remaining 200 tasks are divided into two private test sets.
They are kept confidential to ensure that neither the AI approaches nor their programmers can optimise for them in advance.
One of the private test sets has been used as an undisclosed test set in various programming competitions since 2020, while the second one remains unused and confidential.
In 2024, an additional semi-private test set with 100 newly created tasks was released to evaluate larger AI models that require API access and where the confidentiality of the test set can therefore not be guaranteed \citep{arcprize-2024results}.

Each task consists of a small number of example pairs and one or more test pairs, with each pair consisting of an input and an output grid.
Each grid can have between one and thirty cells in width and height, with the two dimensions being independent of each other.
Each cell can be in one of ten possible states, usually represented by colours for easier interpretation by humans.
In each task, all inputs are manipulated according to a task-specific rule, which results in the corresponding outputs.
For instance, a rule can be that all cells of a certain colour have to be changed to a different colour, or that the input grids have to be mirrored horizontally.
All rules are based only on core knowledge, that is fundamental human beliefs such as the existence of objects or basic algebraic and geometric principles.
To solve a task, the task-specific rule has to be determined by analysing the example pairs and then applied to the test input(s) to generate the test output(s).
A task is only considered solved if the submitted test output corresponds exactly to the correct solution in every single cell state; otherwise the task is considered failed.
Each task is designed so that there is exactly one possible correct solution for each test output \citep[pp. 46-51]{674}.

Following the publication of the article introducing the Abstraction and Reasoning Corpus in 2019, a public competition was held on Kaggle in 2020, where the best approach achieved a 21 \% success rate on the private test set \citep{kaggle-2022}.
In subsequent competitions in 2022 and 2023, hosted by Lab42, the highest score reached was 30 \% \citep{lab42-arcathon2023}.
In the 2024 Kaggle competition, with possible prizes totalling 725,000 USD, the winning approach achieved 53.5 \% \citep{818, kaggle-2024}.
Shortly thereafter, OpenAI's o3 achieved 87.5 \% on the semi-private test set, which is intended to be similar in difficulty to the private test set.
However, o3 was not subject to the computational restrictions imposed by the competitions; instead, its computational costs are estimated to be approximately USD 346,000 \citep{arcprize-2024results}.

Consequently, the question arises as to whether the success of o3 on the ARC-AGI benchmark, which was explicitly designed to test intelligence, is evidence that o3 exhibits intelligence -- or whether ARC-AGI is only of limited suitability for measuring intelligence and other benchmarks are therefore needed to measure progress towards AGI.
Section \ref{sec2} assesses the concept of intelligence and, building on Chollet's distinction between skills and intelligence, introduces a more foundational definition of intelligence, which is aligned with the No Free Lunch theorems.
Section \ref{sec3} analyses the suitability of ARC-AGI as a benchmark for intelligence and for measuring progress towards AGI.
This includes an analysis of the type of problem structure that ARC-AGI tasks represent, as well as the weaknesses that ARC-AGI exhibits and the extent to which these can be overcome.
Section \ref{sec4} outlines a new benchmark for intelligence that is based on the definition of intelligence introduced in the second section and is intended to enable a more comprehensive assessment of intelligence.
Section \ref{sec5} concludes with an evaluation of the performance of OpenAI's o3 on ARC-AGI.

\section{On the Nature of Intelligence} \label{sec2}

To be able to evaluate the occurrence of intelligence, it is important to understand its nature precisely.
Human intelligence is explained by the Cattell-Horn-Carroll theory as an interaction between crystallised intelligence and fluid intelligence \citep[pp. 73-75]{797}:
Crystallised intelligence consists of several broad cognitive abilities, such as reasoning, processing visual information, and remembering information.
Fluid intelligence is a general ability whose performance affects all broad abilities and describes the general cognitive capacity.
In the field of AI, a variety of definitions of intelligence are used \citep{798}, which can be broadly categorised into two groups:
Process-oriented definitions name required abilities such as learning, abstraction, logical thinking, and problem solving.
Result-oriented definitions focus on the outcome and define intelligence as the ability to achieve specific goals; for instance, to adjust to an environment, to create products, or to grasp truths.

To determine whether an AI approach is intelligent, it is usually tested on tasks that fulfil the requirements of the definitions.
In the course of the history of AI, numerous tasks were proposed whose solutions were assumed to require extensive cognitive abilities, and therefore intelligence.
Proposed tasks included playing chess, playing Go, image recognition, translating texts, or creating meaningful texts.
However, when AI approaches were able to solve any of the problems, they were considered not intelligent.
One reason for this is that the methods used by the approaches to solve a task, for example, by trying out a large number of possibilities, are not considered intelligent.
It is also argued that the tasks are not solved by the intelligence of the AI approaches but by the intelligence of the programmers embedded in the approach.
Moreover, it is argued that an approach cannot be intelligent if it can solve a task but fails if the task is modified; a problem that concerns many approaches.
This leads some to conclude that AI approaches are making major progress in terms of performance but not in terms of intelligence (\citealp[pp. 396-404, 421-423, 434]{799}; \citealp[pp. 7-9, 16f]{674}).

\citet[pp. 3-7]{674} explains this contradictory development by the fact that two different interpretations of intelligence are used and that they are not distinguished sufficiently clearly.
The first interpretation understands intelligence as a collection of task-specific skills, as advocated by Darwin and Minsky, for example.
The second interpretation understands intelligence as the ability to create new skills for solving tasks, as advocated by Turing and McCarthy, among others.
Accordingly, while the first interpretation classifies solving tasks known to an AI approach as intelligent, the second interpretation classifies solving tasks hitherto unknown to an approach as intelligent.
\citet[pp. 18-20]{674} argues that the first interpretation of intelligence as task-specific skills is misleading because it does not describe intelligence but only its output:
Skills are specific solutions to specific problems that are created by intelligence but that are not intelligence itself.
In contrast, the second interpretation describes intelligence as a process, as an ability that creates skills.

A further reason in favour of the second interpretation of intelligence is that only that one is suitable for the development of AGI.
This, as skills can be applied to specific tasks for which they were created, i.e. tasks that are known and well-defined, such as mastering games.
But skills cannot be reliably applied to tasks outside the well-defined domain for which they were created:
Skills do not include specifications on how to handle unfamiliar conditions\footnote{
In the field of AI, conditions are often called states.
} that occur outside the well-defined domain.
Everyday tasks from the human domain, which AGI is supposed to solve, often have unfamiliar conditions:
The future development of the world is only partially predictable for humans -- and thus also for skill-based AI approaches created by humans -- and future conditions remain partially unknown.
Accordingly, AGI cannot be realised by a skills-based approach, as it would not be able to handle the constantly arising new, unknown conditions.
Instead, AGI must be able to create new skills to cover the unknown conditions, i.e. AGI must be able to fulfil the second interpretation of intelligence.

The foregoing considerations allow for a more precise definition of skill and intelligence:
A skill is the ability to achieve a specific goal under specific known conditions.
Intelligence is the ability to create new skills that allow to achieve goals under previously unknown conditions.\footnote{
As such, intelligence is also a skill.
It can be considered as a meta-skill that allows to create other skills.
}
Intelligence is not a fixed ability that is only either present or absent, but one that can also be stronger or weaker:
An agent\footnote{
An agent is a system that is able to perform particular actions depending on particular conditions to achieve particular goals.
} is the more intelligent, the more efficiently it can achieve the more diverse goals in the more diverse worlds\footnote{
A world is a system that can have different conditions, some of which are accessible to the agent and some of which may be manipulable by the agent.
Instances of worlds are the universe in which humanity is situated, games such as Go and computer games, or mathematical and logical systems.
} with the less knowledge.
Knowledge is understood pragmatically here: It does not have to be true statements about the worlds, but it includes all information the agent has, including skills.
The exclusion of knowledge in the definition of intelligence entails that only the ability to generate skills, but not skills themselves, falls under intelligence.
The definition thus corresponds to the second interpretation of intelligence discussed by Chollet above and excludes the first interpretation.
Simply put, intelligence describes how well an agent can achieve goals in new, unknown conditions.

The juxtaposition of the application of existing skills on the one hand, and the generation of skills, i.e., intelligence, on the other, reveals a fundamental relationship between the two:
Tasks can be solved either by skills or by intelligence.
This means skills and intelligence can be substituted for each other, provided that all conditions are known.
Intelligence is only necessary to the extent conditions are unknown or skills are not available for other reasons; for example, because skills cannot be provided for all possible known conditions.
The assessment of the degree of intelligence is abstract in that it does not permit a quantitative assessment without further specification of how this is to be carried out.
For example, the assessment does not describe how exactly efficiency or diversity are quantified, or how the individual factors are weighed against each other.
However, the provision of such specifications is not necessary for the further course of the article.
\citet[pp. 27-42]{674}, who provides a measurable definition of intelligence, states that many possible ways of measuring intelligence may be valid.
Which specific quantitative valuation is the best requires further research and may depend on epistemic as well as ontological assumptions.

The No Free Lunch (NFL) theorems show that across all possible optimisation problems any algorithm has the same average performance as every other.
Consequently, there is no algorithm that is better than others at solving all optimisation problems:
If an algorithm performs better than another on one set of optimisation problems, it performs worse than the other on the set of all other optimisation problems (\citealp[pp. 69-71]{801}; \citealp[pp. 4f]{802}).
This can be seen as a counterargument to the formalisation of intelligence:
Intelligence is about solving unknown optimisation problems with above-average performance, but the NFL theorems indicate that there can be no such algorithm.
However, intelligence can only be applied to worlds that have some regularities:
If all conditions of a world were irregular, for example because they were completely random, intelligence could not derive any knowledge about the structure of the world, and skills could not provide any guidance as to which actions are suitable under which circumstances to achieve a particular goal.
This means intelligence is not optimised for all possible optimisation problems, but only for the subset that appears in worlds with regularities \citep[cf.][pp. 402f]{799}.
Consequently, it is possible to find an algorithm that performs better than others on this subset of problems -- and worse on the remaining optimisation problems of completely irregular worlds.

For an algorithm to be better than others on a subset of optimisation problems, the characteristics of the subset must be incorporated into the algorithm \citep[pp. 71f]{801}.
In the case of intelligence, the algorithm has to be optimised with regard to regularities \citep[cf.][pp. 1300f]{766}.
The considered regularities are thereby not a necessary truth of the worlds, but an assumption.
The formalisation of intelligence therefore faces a dilemma when it comes to determining to what extent regularities -- and possible other assumptions -- should be considered:
The more assumptions are considered, the smaller the subset of optimisation problems covered and the more efficient the algorithm, everything else being equal.
However, the more assumptions are considered, the higher the chance that they do not correspond to the worlds to which the algorithm is applied to, and its performance decreases accordingly.

\section{Suitability of ARC-AGI as a Benchmark for AGI} \label{sec3}

Presenting o3's achievement on ARC-AGI, \citet{arcprize-o3} concludes: 
'ARC-AGI serves as a critical benchmark for detecting such breakthroughs, highlighting generalization power in a way that saturated or less demanding benchmarks cannot. However, it is important to note that ARC-AGI is not an acid test for AGI.'
\citet{arcprize-o3} therefore proposes to develop a new version of ARC-AGI, a 'next-gen, enduring AGI benchmark' in the same format.
This poses the question to what extent ARC-AGI in its current form is suitable for measuring the capacity for broad generalisation, and to what extent and in what way it is possible to develop it further in the same format.

ARC-AGI is different compared to most other benchmarks in that it is not designed to measure how good AI approaches are in a particular skill, but instead in their ability to generalise \citep[pp. 4, 53f]{674}.
To accomplish this, the test set is kept secret, the tasks are designed to be diverse, and each task has only a few examples from which to generalise.
By limiting the required knowledge to core knowledge, the emphasis is not on the use of existing knowledge, but on the ability to abstract and reason.\footnote{
\citet[pp. 47-50]{674} considers the core knowledge used to be explicitly described and complete.
Yet it appears to be incomplete; for example, Boolean functions such as AND, OR, NOT are not mentioned but occur in several ARC-AGI tasks.
In addition, it is unclear to what extent concepts that can be derived from described concepts are also considered valid.
For example, from the included concept of addition, the concept of multiplication can be derived -- a concept which is also used in ARC-AGI tasks.
Equally, the concept of division and, with the help of this, the concept of prime numbers could be derived.
}
All these factors together place the focus on fulfilling the second interpretation of intelligence, i.e. the capacity to develop solutions for new, previously unknown tasks by means of generalisation.
The minimalistic design of ARC-AGI tasks as simple, coloured grids, whose transformation can be described using core knowledge only, allow for easy development and testing of new AI approaches.

However, the minimalist and specific design of the ARC-AGI tasks also represents a very specific problem structure for the following two reasons:
First, to solve an ARC-AGI task, it is necessary to determine the most simple transformation rule that describes the changes between the input and output example grids.
The determined transformation rule then has to be applied to the test input(s) to generate the test output(s).
Each transformation rule can be described by a combination of core knowledge.
The entire core knowledge can be represented by a finite and small set of operations that determine certain properties of the grids or apply certain changes to them.\footnote{
An example of the implementation of core knowledge in the form of a finite and small set of operations is provided by \citet{git-dsl-hodel}.
}
Although for each task a different transformation rule has to be determined and the large number of possible combinations of core knowledge allows a greater variety, the underlying problem structure is always the same:
From the existing, small and finite set of potential core knowledge operations, those that together correctly describe the transformation must be selected and combined together.
Second, ARC-AGI tasks represent a very specific problem structure as each task is required to have a single correct solution, i.e. there is exactly one correct output grid for each input grid.
This makes it possible to test the correctness of possible transformation rules:
A transformation rule is correct exactly then when it determines in every example pair for the input grid the correct corresponding output grid.
Consequently, since the example pairs are given, it is possible to check whether a transformation rule is correct or not before submitting a solution.

Both characteristics of the problem structure of ARC-AGI tasks allow the solution process to be considerably simplified in the following two regards:
First, many problems require a solution process that can be described by a combination of exploration and exploitation \citep[cf.][]{827}.
Exploration describes the process of representing a task in a form that allows a solution to be found; for example, to find the best route to a distant location, the task can be framed as a cost optimisation problem.
Exploitation describes the process of finding the optimal solution within the representation of the task; for instance, by comparing the costs in money and time for different travel routes.
While the exploitation of a problem representation can often be considered as an optimisation problem, exploration is considerably more challenging as it requires a suitable framing of the problem, i.e. the creation of a functional representation.
This requires the identification of the relevant aspects that need to be considered as well as the creation of a model that represents the relationships between them (\citealp[cf.][pp. 7-9]{711}; \citealp[sect. 5]{829}).
The problem structure of ARC-AGI already implies a specific representation of the problem:
For every task, the simplest possible transformation rule has to be found, which has to be composed of given core knowledge operations.
Consequently, AI approaches can solve ARC-AGI by relying only on the exploitation of a given problem representation, without the need for exploration, i.e. the creation of a suitable problem representation, beforehand.

Second, the solution process can be considerably simplified in the following way:
Since the correctness of a transformation rule can be tested on the example pairs, ARC-AGI allows, within the limited computational resources, for unrestricted trialling of possible transformation rules.
While unrestricted trialling works well for ARC-AGI and other mathematical problems \citep[cf.][]{828}, such an approach does not work for many other types of problems:
For many problems, especially in the physical world, but also in many other domains, one often has only one or at most a few attempts to check whether a solution is correct or not.
For instance, pressing the wrong combination of buttons on a coffee machine will spoil the drink, and driving a car incorrectly can lead to a serious accident.
There are ways to circumvent such problems, e.g. a robot, before grasping a cup of coffee, can simulate the grasping process in a virtual physical environment and thereby find a way to successfully hold the cup.
However, this only works for domains that are sufficiently known so that all relevant aspects can be considered in the simulation -- in other words, it only works for tasks whose conditions are already known, i.e. they must be realised as a skill.
For tasks whose conditions are not known, this method does not work.
Equally, it does not work for tasks that cannot be simulated for other reasons; for example, because tasks are too complex, require too many computing resources, or actions must be performed faster than their simulations could be carried out.

In addition to the major weakness that the problem structure of ARC-AGI allows a much simpler solution finding process than many other problems, there are several other weaknesses.
For example, while the ARC-AGI benchmark limits the computing resources allowed for the solution finding process, it does not reflect the cost of training the approaches beforehand.
Yet, training can significantly improve the score of an approach; for instance, during the 2024 ARC-AGI competition, some participants trained their approaches on a large number of artificially generated ARC-AGI tasks.
This is particularly of concern as training is a sign that an approach is not based on intelligence but on skills -- whereas ARC-AGI intends to measure the former.
Furthermore, although the test set is kept private, competition participants had the possibility to run their approaches several hundred times on it, allowing them to probe it and optimise their approach specifically for it.
In summary, although ARC-AGI was created with the intention of being solvable only by broad generalisation, it has several features that make it vulnerable to additional solution methods:
The specific type of problem structure that ARC-AGI tasks represent is not only much easier to solve than many other types of problems, it also represents a very small subset of the huge diversity of possible problems, which supports the application of skill-based approaches.
The possibility of massively trialling solutions allows approaches to test a large number of possible, low-quality solutions that may not be obtained by intelligence but, for example, by guided guessing.
This, prior training, and probing of the test set allows ARC-AGI to be solved not by means of broad generalisation, but by skills-based approaches that are specifically optimised for ARC-AGI.

This raises the question of whether ARC-AGI can be improved to overcome its weaknesses while retaining the same format.
Some of the issues can be overcome, for instance a new private test set can be used for which probing can be prohibited.
However, the specific type of problem structure ARC-AGI represents is an inherent aspect of the current format and cannot be overcome by minor adjustments.
Instead, addressing this aspect requires an entirely new type of benchmark that allows for other types of problems which represent a much greater diversity, require exploitation, and do not allow for massive trialling of possible solutions.
Nevertheless, although ARC-AGI does not appear to be a sufficiently suitable intelligence benchmark for the future, it has to be concluded with \citet{arcprize-o3}:
'It's a research tool designed to focus attention on the most challenging unsolved problems in AI, a role it has fulfilled well over the past five years.'

\section{Towards a New Benchmark for Intelligence} \label{sec4}

The limitations of ARC-AGI lead to the question of how a benchmark can be designed that can be used to measure intelligence and progress towards AGI.
A noticeable characteristic of the ARC-AGI benchmark is that it appears to be subject to Goodhart's Law:
'When a measure becomes a target, it ceases to be a good measure.'
This was evident in many places in the various ARC-AGI competitions.
Instead of the approaches being developed towards AGI, they were developed to achieve the highest possible score on the test set:
Core knowledge representations were optimised, ARC-AGI-specific skills were improved using artificially generated training data, and some participants submitted hundreds of approaches to probe and optimise for the test set.
In order to avoid this, a future benchmark should have the greatest possible correspondence between the measure and the target, i.e. the development of (artificial) intelligence.
To express it more directly: The best benchmark for intelligence is intelligence itself.

Consequently, an ideal benchmark should rate an agent as the more intelligent, the more efficiently it can achieve the more diverse goals in the more diverse worlds with the less knowledge.
While human intelligence tests tend in this direction, they do so only to a very limited extent:
They use time as a measure of efficiency, age as a measure of existing knowledge, and test various goals in various domains.
Altogether, however, the tests are neither precise nor very diverse, and to a large extent they measure skills instead of intelligence.
This is not least because human intelligence tests are designed to predict human performance in the human domain.

In contrast, AGI, as a formal system, is neither bound to the human domain nor to the physical world.
Instead, AGI can be given any type of goal which it has to fulfil in any type of world.
These can be worlds that humans can access and understand, such as games, but also completely arbitrary worlds that can be simulated.
For example, AGI can be situated in a simulation of a universe that is fundamentally different from ours.
The universe could have more or fewer dimensions than ours, and fundamental aspects such as the laws of physics or the principle of causality could be altered.
An AGI approach to be evaluated can be given any type of embodiment in such a universe -- or none -- as well as any type of goal to fulfil.
The better and more efficiently it achieves the goals, the more it is rated intelligent.

A concrete example of the outlined benchmark could look as follows:
All the approaches to be tested have to solve tasks in ten different worlds; nothing is known about the worlds in advance.
One such world could be for example a simulation of Mars.
In this world, the AI approach is embodied in a Mars robot, for which it must first develop an understanding.
The approach must then fulfil the task of building an accommodation for astronauts, for which it must develop an understanding of the physical conditions on Mars.
Another world could be, for example, the simulation of a gas planet in a four-dimensional universe in which the approach has to produce a specific chemical element using an alien body.
A further world could be simulated by a digital strategy game in which the AI approach has to win against human players.
This requires the approach to develop an understanding of the world of the computer game and also a differentiated picture of agentness, which allows it to predict and anticipate the actions of the other players.
A next world could be one in which the AI approach has no possibility of manipulating the world, e.g. through a body, but still has to make predictions about future states by analysing the previous states of the world.
Such a world could be as different as predicting the future development of a quantum world or a habitat, or how well a newly launched product will sell based on available market data.
In addition to these described worlds, the AI approaches are tested in several further worlds to ensure a greater diversity of the test set.
An approach is never tested multiple times in the same world to prevent it from using previously acquired knowledge instead of intelligence.
The approaches are evaluated according to how efficiently they fulfil the specified goals, for example how much time they need to do so and how comprehensively they fulfil them.
The more efficiently an AI approach fulfils the more goals in the more worlds, the more intelligent it is rated.
As the tasks illustrated here require intelligence at least on a human level, it is possible to start with the simulation of simpler worlds and goals.
This makes it possible to identify AI approaches that only have a lower level of intelligence but are nevertheless more successful than others.
As the level of intelligence of AI approaches increases, the worlds and the tasks to be performed in them can be made increasingly difficult.

The worlds tested in the benchmark may no longer be accessible or understandable for humans, as humans are limited by their capabilities and bound by their skills.
Nevertheless, this does not pose an issue for AGI, on the contrary.
For an AGI approach to be successful in such arbitrary worlds, a programmer must focus exclusively on the implementation of intelligence.
Any implementation of world-specific skills, such as human core knowledge or a specific understanding of causality, would not only be pointless but even detrimental -- the approach would be impaired by the skills if they are not feasible in the respective world.
A benchmark based on the assessment of the degree of intelligence is not perfect in that the types of worlds that can be generated still have a certain degree of conformity:
they all need to be formalisable, executable with current computing limitations, and are limited by human imagination.
Theoretically, this conformity allows the realisation of a skill other than intelligence that is tailored to these worlds:
With enough training, an approach can be successful in these worlds to a limited extent, just as LLM-based systems are successful not through reasoning but through knowledge in some parts of the human domain.
However, not only will there be a much greater diversity to manage with much less available knowledge, but the benchmark could also require that the training time of the approaches is taken into account and that the approaches have to be of a certain compactness.

\citet[pp. 20-24]{674} argues that AGI should be benchmarked against human intelligence, as intelligence would need to be tied to a precisely defined area of application and only those areas relevant to humans can be accessed.
Yet this is not necessary, as the above assessment of the degree of intelligence shows: any goal can be specified in any world -- only the degree of goal fulfilment has to be measurable.
Instead, focussing on human areas of application harbours several risks:
First, the development of AGI may focus on skills instead of intelligence, as has often happened in the history of AI.
Second, it would limit AGI too much to problems from the human domain, although problems from other domains can also be of interest to humans.
For example, AGI can discover the world of bacteria or of deep space, both domains in which human skills are likely to be of limited use, and make them comprehensible to humans.
Lastly, there is a risk that human assumptions will be taken too much into account.
For instance, \citet[pp. 47-50]{674} refers to core knowledge that represents very basic beliefs from a human perspective, but which are at the same time very specific and convey a very particular view of the world. 
The same applies, for example, to classical physics, which is regarded as a fundamental theory of our universe.
However, it can only be applied to a limited range of physical phenomena and contradicts quantum physics; both are indications that the theory does not represent the true nature of the universe, but is a pragmatic model that fulfils human needs.
Providing classical physics as axioms to AGI would therefore limit its capabilities, as it would be constrained by these flawed beliefs.

In conclusion, to measure intelligence reliably, a benchmark should be created that provides AGI approaches randomly generated worlds that are as diverse as possible and whose only commonality is that each of it has some regularities.
All AGI approaches are measured by how efficiently they can achieve various, as different as possible goals in these environments.
To this end, the approaches must identify the often hidden regularities and utilise them to determine the best available measures to achieve the goals \citep[cf.][]{829}.
Many more details of the benchmark have to be specified and it has to be implemented in practice.\footnote{
A possible environment for implementing an initial version of the benchmark could be a modified version of \citet{git-genesis}, for example.
}
The benchmark outlined here should be universal, i.e. it should remain valid regardless of the approaches tested and their operating characteristics.
Nonetheless, as with ARC-AGI, it is possible that adaptations to new developments will be necessary once the benchmark is applied, as benchmarking AGI involves the assessment of a moving target.
This requires a continuous understanding of new approaches and their impact on the validity of benchmark tests in order to create a benchmark that addresses recognised shortcomings on the path towards AGI.
Nevertheless, with the measure of intelligence so close to the target, the benchmark outlined here should require only minor adjustments and generally strongly support the development of new approaches that represent progress towards AGI.

\section{Conclusion} \label{sec5}

The analysis of ARC-AGI has shown that it cannot serve as a benchmark for general intelligence and thus as a measure of progress towards AGI.
Its simplicity, which makes it ideal for developing and testing new approaches, brings with it several weaknesses.
Most importantly, the tasks have a very specific problem structure that allows the tasks to be solved exploiting a known problem representation without having to create one first, although this is often the more difficult part of solving problems.
In addition, it enables a massive trialling of possible solutions, allowing a high score to be achieved by the massive generation of low quality solutions.

Based on these insights, it is possible to analyse o3's score of 87.5 \% on ARC-AGI in more detail.
To achieve the score, o3 incurred estimated computing costs of USD 346,000 -- equivalent to USD 3,460 per task.
A low-compute version of o3, achieving 75.7 \% on the semi-private test set, incurred computing costs of USD 2,012 -- equivalent to USD 20 per attempted task \citep{arcprize-o3}.
'The reason why solving a single ARC-AGI task can end up taking up tens of millions of tokens and cost thousands of dollars is because this search process has to explore an enormous number of paths through program space' \citep{arcprize-o3}.
Although this method can achieve a high score, given sufficient computing power, it cannot be regarded as very efficient.\footnote{
The original ARC-AGI benchmark used in the competitions is subject to strict limitations in terms of computing power and would therefore not allow such a high score using this method.
The result of o3 is only possible because the computing restrictions were waived for its testing.
}
Furthermore, this method is only suitable for a very specific type of problem, but not for most problems in the physical world or in the human domain, where massive testing of solutions in advance is not possible.
The method also does not correspond well with the original intention of ARC-AGI: the development of new AI approaches that can reliably abstract and reason, and thus can determine the correct solution on the first or at least the first few attempts.
While LLM-based systems appear to have some capacity for abstraction and reasoning -- both processes considered fundamental to intelligence -- they do not appear to perform them reliably \citep{826, 825, 824, 785, 819, 783}.
Instead, they seem to rely to a greater extent on memorisation, i.e. the application of skills \citep{784, 823, 820, 821, 822, 786}.
Overall, o3's performance on ARC-AGI is not due to intelligence but due to the application of knowledge and computing resources that together enable an effective search in the given space of possible solutions.

This raises the question of how approaches can be developed that are centred more on intelligence.
Building on the above, intelligence is not about how much data is processed, or how extensively it is processed, but about how it is processed.
In other words, progress towards AGI requires a shift from datasets and computing resources towards the algorithm itself.
Abstraction and reasoning, the two core components of ARC-AGI, appear to be important aspects of intelligence and could provide further insights into how an algorithm of intelligence could be designed.
Much more research is required; for example, reasoning can be categorised into deduction, induction \citep[cf.][]{661} and abduction \citep{711}, but many details are still unclear.
It is therefore hoped that the benchmark outlined in this article contributes to further exploration of this direction of research and incentivises the development of new AGI approaches that focus on intelligence rather than skills.

\bibliography{sn-article}

\end{document}